\documentclass[sigconf]{acmart}

\usepackage{booktabs} 
\usepackage{multirow}
\usepackage{amsfonts}  
\usepackage{amsmath}  
\usepackage{mathrsfs}  
\usepackage{bm}
\usepackage{cases}  
\usepackage{algorithm}
\usepackage{algorithmic}

\newtheorem{myDef}{Definition}
\usepackage{makecell}

\AtBeginDocument{%
  }

\copyrightyear{2024}
\acmYear{2024}
\setcopyright{rightsretained}
\acmConference[CIKM '24]{Proceedings of the 33rd ACM International Conference on Information and Knowledge Management}{October 21--25, 2024}{Boise, ID, USA}
\acmBooktitle{Proceedings of the 33rd ACM International Conference on Information and Knowledge Management (CIKM '24), October 21--25, 2024, Boise, ID, USA}
\acmDOI{10.1145/3627673.3679845}
\acmISBN{979-8-4007-0436-9/24/10}

\makeatletter
\gdef\@copyrightpermission{
	\begin{minipage}{0.3\columnwidth}
		\href{https://creativecommons.org/licenses/by/4.0/}{\includegraphics[width=0.90\textwidth]{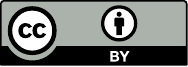}}
	\end{minipage}\hfill
	\begin{minipage}{0.7\columnwidth}
		\href{https://creativecommons.org/licenses/by/4.0/}{This work is licensed under a Creative Commons Attribution International 4.0 License.}
	\end{minipage}
	\vspace{5pt}
}
\makeatother

\begin{document}

\title{Look Globally and Reason: Two-stage Path Reasoning over Sparse Knowledge Graphs}


\author{Saiping Guan}
\affiliation{%
	\institution{Key Laboratory of Network Data Science and Technology, Institute of Computing Technology, Chinese Academy of Sciences}
	\city{Beijing}
	\country{China}}
\affiliation{%
	\institution{University of Chinese Academy of Sciences}
	\city{Beijing}
	\country{China}}
\email{guansaiping@ict.ac.cn}

\author{Jiyao Wei}
\affiliation{%
	\institution{Key Laboratory of Network Data Science and Technology, Institute of Computing Technology, Chinese Academy of Sciences}
	\city{Beijing}
	\country{China}}
\affiliation{%
	\institution{University of Chinese Academy of Sciences}
	\city{Beijing}
	\country{China}}
\email{weijiyao20z@ict.ac.cn}

\author{Xiaolong Jin}
\affiliation{%
	\institution{Key Laboratory of Network Data Science and Technology, Institute of Computing Technology, Chinese Academy of Sciences}
	\city{Beijing}
	\country{China}}
\affiliation{%
	\institution{University of Chinese Academy of Sciences}
	\city{Beijing}
	\country{China}}
\email{jinxiaolong@ict.ac.cn}

\author{Jiafeng Guo}
\affiliation{%
	\institution{Key Laboratory of Network Data Science and Technology, Institute of Computing Technology, Chinese Academy of Sciences}
	\city{Beijing}
	\country{China}}
\affiliation{%
	\institution{University of Chinese Academy of Sciences}
	\city{Beijing}
	\country{China}}
\email{guojiafeng@ict.ac.cn}

\author{Xueqi Cheng}
\affiliation{%
	\institution{Key Laboratory of Network Data Science and Technology, Institute of Computing Technology, Chinese Academy of Sciences}
	\city{Beijing}
	\country{China}}
\affiliation{%
	\institution{University of Chinese Academy of Sciences}
	\city{Beijing}
	\country{China}}
\email{cxq@ict.ac.cn}

\renewcommand{\shortauthors}{Saiping Guan, Jiyao Wei, Xiaolong Jin, Jiafeng Guo, \& Xueqi Cheng}

\begin{abstract}
  Sparse Knowledge Graphs (KGs), frequently encountered in real-world applications, contain fewer facts in the form of (head entity, relation, tail entity) compared to more populated KGs. The sparse KG completion task, which reasons answers for given queries in the form of (head entity, relation, ?) for sparse KGs, is particularly challenging due to the necessity of reasoning missing facts based on limited facts. Path-based models, known for excellent explainability, are often employed for this task. However, existing path-based models typically rely on external models to fill in missing facts and subsequently perform path reasoning. This approach introduces unexplainable factors or necessitates meticulous rule design. In light of this, this paper proposes an alternative approach by looking inward instead of seeking external assistance. We introduce a two-stage path reasoning model called LoGRe (Look Globally and Reason) over sparse KGs. LoGRe constructs a relation-path reasoning schema by globally analyzing the training data to alleviate the sparseness problem. Based on this schema, LoGRe then aggregates paths to reason out answers. Experimental results on five benchmark sparse KG datasets demonstrate the effectiveness of the proposed LoGRe model.
\end{abstract}




\maketitle

\section{Introduction}
\label{sec:introduction}
Sparse Knowledge Graphs (KGs) are a special type of KGs, containing fewer facts in the form of (head entity, relation, tail entity) compared to more populated KGs. They are frequently encountered in real-world applications, such as location recommendation, fund recommendation, and fraud detection~\cite{POI_recommand2019,POI_recommand2020,CIKM-App-Recommend,fraud_prediction2023}. The sparseness of sparse KGs impedes the performance improvement of such applications. Hence, sparse KG completion, formalized as reasoning answers for given queries in the form of (head entity, relation, ?) for sparse KGs, is a crucial and urgent task.

The sparse KG completion task is challenging, as it requires reasoning missing facts based on limited available facts. Thus, the performance of existing models for KG completion drops significantly when applying them to sparse KGs~\cite{pujara-2017-sparsity,DacKGR,SKGC2023-KRACL,SKGC2023-RelaGraph}. To address this, researchers have specially extended these models. They can be classified into embedding-based, rule-based, and path-based models. Before devising a score function to determine the likelihood of a candidate fact or candidate tail entity, embedding-based models carefully design the embedding network to incorporate additional information, such as neighboring facts~\cite{SKGC2023-RelaGraph,SKGC2023-KRACL}, external textual information~\cite{SKGC2022-BERT,VEM2L}, and external KGs~\cite{SKG2023-Application}, which lacks explainability. Rule-based models utilize embedding-based models (and transfer learning) to derive rules from sparse KGs, which are then employed to reason out answers~\cite{SKGC2020-Structure,Rule-SKGC-2019,Rule-SKGC-2023}. Path-based models typically use embedding-based or rule-based external models to fill in missing facts and then perform path reasoning to obtain answers~\cite{DacKGR,SparKGR,Hi-KnowE}. However, these models either rely on unexplainable embedding-based reasoning facts or require carefully designed rules.

With these considerations in mind, this paper follows the line of path-based models for sparse KG completion but emphasizes internal exploration rather than relying on external assistance. The primary challenge lies in how to effectively address the sparseness problem. It is difficult to reason missing facts from an individual standpoint. However, from a global perspective, it is possible to identify paths leading to answers. For instance, in a sparse KG, person $A$ may possess relations $r_1$ and $r_2$ while missing others; person $B$ may only have relations $r_1$ and $r_3$, etc. We can group the relations of the same entity type together and obtain relatively complete relations associated with the corresponding entities. We can then extract reasoning paths for each relation and assign a score to each path. We can also identify relations shared across various entity type groups and combine them to calculate the overall scores of their paths. This process yields a pool of relation-path reasoning elements called relation-path reasoning schema in this paper, with elements categorized by entity types (i.e., type-specific relation-path reasoning schema) and a shared group (i.e., cross-type relation-path reasoning schema). Based on this schema, we aggregate the endpoints of the related paths to reason out answers for given queries in the form of (head entity, relation, ?). The underlying rationale is that individual indeterministic paths cannot be relied upon in isolation, but if they all converge towards the same answer, a sensible result can be inferred.

Based on this simple idea, we propose a two-stage path reasoning model for sparse KG completion. It Looks Globally to construct a relation-path reasoning schema and then performs path Reasoning, thus called LoGRe. Generally, our main contributions are as follows:
\begin{itemize}
	\item For the first time in the line of path-based sparse KG completion models, we propose to look inward rather than relying on external assistance.
	\item We introduce a two-stage path reasoning model, LoGRe, for sparse KG completion, which constructs a relation-path reasoning schema from a global perspective and then aggregates paths to reason out answers.
	\item Experimental results on five benchmark datasets and further analyses validate the superiority of LoGRe.
\end{itemize}

\section{Problem Statement}
\label{sec:problem_statement}

\begin{myDef}
	\textbf{KG} is a graph of entities $E$ and their relations $R$ with facts represented as $(h,r,t)$, where $\{h,t\}\in E$ and $r\in R$, indicating the head entity $h$ has the relation $r$ with the tail entity $t$.
	\label{def:KG}
\end{myDef}

\begin{myDef}
	\textbf{Sparse KG} is a special type of KG, where entities are less connected, containing fewer facts compared to a regular KG. 
	\label{def:SKG}
\end{myDef}
Sparse KG is a relative concept compared to KG. As introduced in DacKGR~\cite{DacKGR}, a KG is considered dense or normal if the average degree of its entities exceeds a certain threshold; otherwise, it is considered sparse. There is no common consensus on the precise value of this threshold.

\begin{myDef}
	\textbf{Sparse KG completion} is the task of reasoning missing facts based on limited facts available in a sparse KG. For example, given a query $(h, r, ?)$, it is to determine answers\footnote{The query $(?, r, t)$ can be transformed to $(h, r, ?)$ by introducing inverse relations. We follow the line of path-based models focusing on $(h, r, ?)$~\cite{DacKGR,SparKGR}.}, also called correct tail entities in this paper.
	\label{def:SKGC}
\end{myDef}

\section{Related Works}
\label{sec:related_works}
This paper handles sparse KG completion. Its literature can be divided into embedding-based, rule-based, and path-based models. 

\subsection{Embedding-based Models}
Embedding-based KG completion models learn embeddings of entities and relations, and devise a score function to determine the likelihood of a candidate fact or candidate tail entity through vector operation. For example, TuckER~\cite{TuckER} utilize multiplication operations. More complexly, ConvE~\cite{ConvE} reshapes and concatenates the embeddings of the head entity and relation to form a matrix, where a 2D convolution operation is applied to predict the tail entity. InteractE~\cite{InteractE} further enhances the interaction between the head entity and relation through feature permutation, feature reshaping, and circular convolution. NBFNet~\cite{NBFNet} adopts a graph neural network to predict the conditional likelihood of the tail entity.

To address the sparseness problem of sparse KGs, researchers incorporate additional information into their models. RelaGraph~\cite{SKGC2023-RelaGraph} extends neighborhood information during entity encoding and incorporates neighborhood relations to mine deeper graph structure information. KRACL~\cite{SKGC2023-KRACL} introduces a knowledge relational attention network to integrate neighboring facts and proposes a knowledge contrastive loss that considers more negative samples. \citet{SKGC2022-BERT} fine-tuned pretrained embeddings of textual information based on the sparse KG to compensate for the lack of structure information. \citet{VEM2L} trained existing structure-based and text-based models, and fused knowledge acquired by them. \citet{SKG2023-Application} even linked to external KGs to introduce additional entity information and employed graph convolutional neural networks to aggregate the information.

\begin{figure*}[!htb]
	\centering
	\includegraphics[width=5.6in]{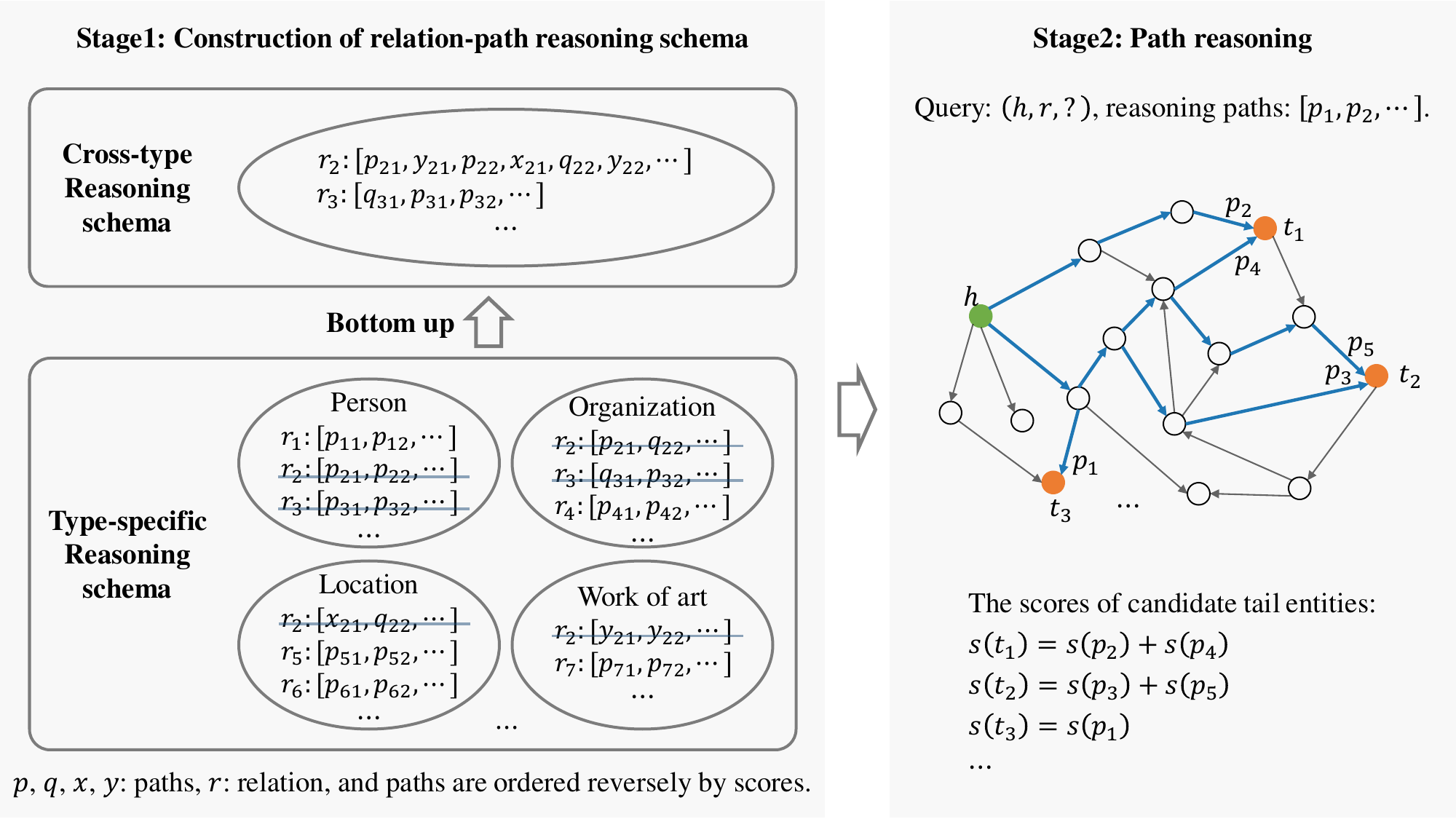}  
	\caption{The framework of the proposed LoGRe model.}
	\label{fig:model_framework}
\end{figure*}
\subsection{Rule-based Models}
Rule-based KG completion models deduce logical rules based on the graph structure of KG and use these rules to reason out answers for given queries. For example, NTP~\cite{NTP} introduces neural networks for end-to-end differentiable proving of queries to KG, where rules are induced using gradient descent. RlvlR~\cite{RlvlR} proposes a target relation-oriented sampling method and a matrix operations-based rule evaluation mechanism to mine massive rules efficiently. RuLES~\cite{RuLES} iteratively extends rules induced from a KG by relying on the feedback from a precomputed embedding model over the KG and external information, including text corpora. AnyBURL~\cite{AnyBURL} introduces Reinforcement Learning (RL) to guide the rule mining process. Based on the rules learned from AnyBURL~\cite{AnyBURL}, \citet{Rule-KGC-2024} further proposed a supervised approach to learn a reweighted confidence value for each rule given a specific aggregation function.

To mitigate the impact of missing facts in sparse KGs on rule mining, researchers seek assistance from other models. \citet{SKGC2020-Structure} proposed an iterative framework of rule learning and embedding, where rules are inferred and scored based on the initial embeddings of entities and relations. New facts are then inferred from rules and embedded into the model. \citet{Rule-SKGC-2019} employed transfer learning to address the sparseness issue. They introduced embedding similarities of entities, relations, and rules to automatically select the most relevant source KGs and rules for transfer. \citet{Rule-SKGC-2023} further constructed mappings between the source KG and the target KG to transfer rules from the source KG to the target KG.

\subsection{Path-based Models}
Path-based KG completion models perform path exploration on KG, starting from the given head entity, to reach answers with the guidance of the given relation. For example, \citet{pr_cbr1,pr_cbr2} applied paths from similar entities of the given head entity to reason out tail entities. RuleGuider~\cite{RuleGuider} leverages high-quality rules to provide reward supervision for RL agents. In contrast, CPL~\cite{CPL} trains two collaborative RL agents jointly: a fact extractor and a multi-hop reasoner. The extractor generates facts from corpora to enrich KG, while the reasoner offers feedback to the extractor and guides it towards promoting facts helpful for the reasoning. 

To address the challenges in selecting correct paths due to the lack of facts in sparse KGs, researchers usually use external models to fill in missing facts before path reasoning. DacKGR~\cite{DacKGR} dynamically adds edges based on embedding-based models as additional actions during the path search in the RL framework. SparKGR~\cite{SparKGR} employs rule guidance to complete missing paths, augmenting the action space for the RL agent. Moreover, it incorporates global information from KG through iterative optimization of rule induction and fact reasoning to guide the exploration of the RL agent. Whereas, SQUIRE~\cite{SQUIRE}, SelfHier~\cite{Cold_start}, and DT4KGR~\cite{DT4KGR} learn to generate paths and answers, and apply AnyBURL~\cite{AnyBURL} to find evidential paths for model training. Hi-KnowE~\cite{Hi-KnowE} conducts hierarchical RL, with high-level policy to reason relations and low-level policy to obtain tail entities. The relation action space is enlarged under rule guidance from AnyBURL~\cite{AnyBURL} and the top results from ConvE~\cite{ConvE} are used to refine the entity action space.

\subsection{Summarization and Comparison}
In general, there have been numerous studies for KG completion, with fewer and more recent studies specializing in sparse KGs. Among them, path-based models are anticipated to offer superior explainability compared to embedding-based models, while obviating the need for meticulously designed rules compared to rule-based models. However, existing path-based sparse KG completion models necessitate external assistance from other models to tackle the sparseness problem. They introduce unexplainable embeddings or rely on high-quality rules, deviating from the original intention of path-based models. Thus, this paper proposes an alternative approach by internally addressing the sparseness problem instead of seeking external assistance.

\begin{figure*}[!htb]
	\centering
	\includegraphics[width=6.95in]{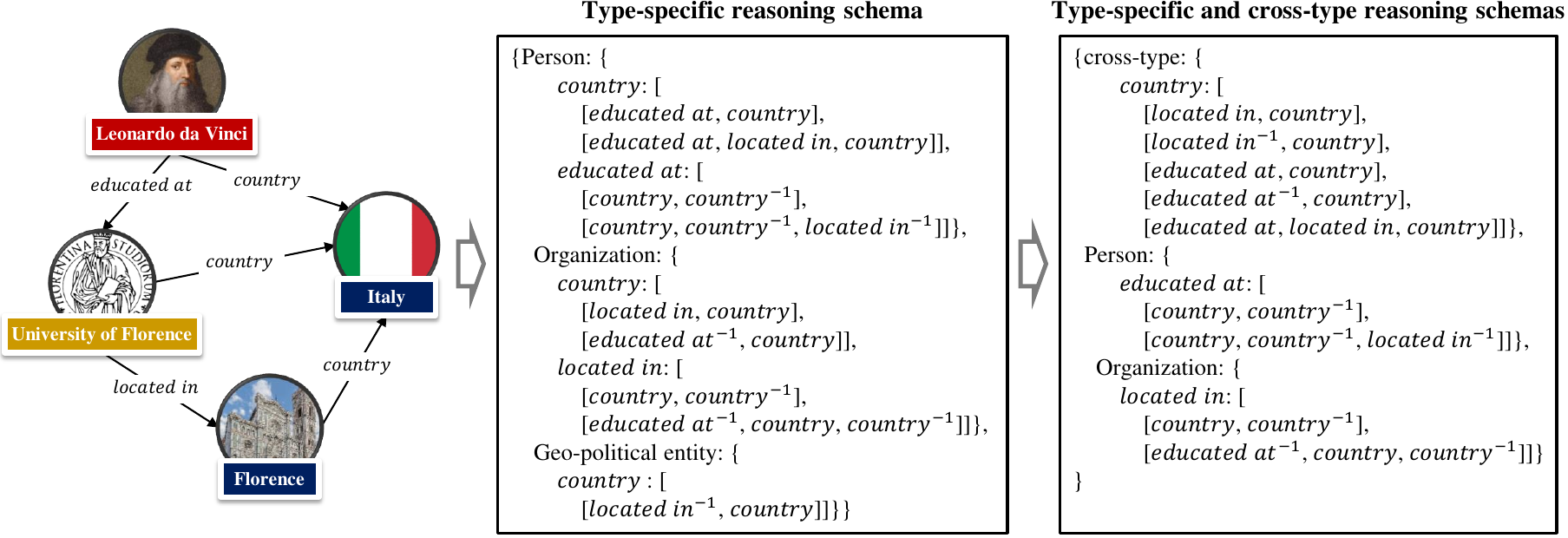}  
	\caption{A small example of the construction process for the relation-path reasoning schema. Dark red, yellow, and blue represent the entities of type person, organization, and geo-political entity, respectively.}
	\label{fig:model_example}
\end{figure*}
\section{The Proposed LoGRe Model}
\label{sec:proposed_model}
We propose a two-stage model called LoGRe to look inward without reliance on external models. As depicted in Figure~\ref{fig:model_framework}, it looks globally to construct a relation-path reasoning schema from the training data and then aggregates paths to reason.

\subsection{Stage1: Construction of Relation-path Reasoning Schema}
This stage globally analyzes the training data to obtain type-specific relation-path reasoning schema (referred to as type-specific reasoning schema), and finally relation-path reasoning schema, which encompasses type-specific and cross-type ones. For clarity, we present a small example of this stage in Figure~\ref{fig:model_example}. Now, Let us elaborate on this stage.

\subsubsection{Type-specific Reasoning Schema}
\textbf{The Construction Process.}
Type-specific relation-path reasoning schema is constructed as follows:
\begin{itemize}
	\item For each entity $e\in E$, we randomly collect a maximum number of $N_{path}$ paths\footnote{If the number of paths for an entity is less than $N_{path}$, the number of collected paths will also be less than $N_{path}$.} from the training data to construct the entity-path dictionary $D$. To avoid the time-consuming collection of excessively long paths that may not contribute to the reasoning, the hop of each path is restricted to a maximum of $N_{hop}$.
	\item Entities are divided into groups based on their types, such as person, organization, and location.
	\item For each group of index $i$, we gather the relation set $R_i$ possessed by its entities $E_i$.
	\item For each relation $r_j\in R_i$, we retrieve the specific paths from the entity-path dictionary $D$ by the entities in $E_i$ that have $r_j$. These paths should lead to the correct tail entities in the training data.~\label{item_condition}
\end{itemize}
Thus, the relations and their corresponding paths constitute an initial type-specific relation-path reasoning schema. As illustrated in the left part of Figure~\ref{fig:model_framework} and Figure~\ref{fig:model_example}, its relation-path reasoning elements grouped by entity types, are in the form of:
\begin{equation}
	r_j: [p_{j1}, p_{j2}, \cdots],
\end{equation}
where $p_{jk}$ ($k=1,2,\cdots$) is the $k$-th path of relation $r_j$ in the group, which satisfies the two conditions mentioned above: At least one entity in the group has $r_j$; path $p_{jk}$ is among the collected paths of these entities in $D$, starting from them and reaching the correct tail entities of $r_j$. Path $p_{jk}$ is a list of relations of the corresponding exploration in the sparse KG, i.e.,
\begin{equation}
	p_{jk} = [r_{jk1},r_{jk2},\cdots],
\end{equation}
where $r_{jkl}$ ($l=1,2,\cdots$) is the $l$-th relation in $p_{jk}$.

\textbf{Path Score Calculation.}
To facilitate path reasoning, we calculate a score for each path. For sparse KGs, where facts and certain paths are missing, we think path precision is crucial. Thus, the score $s(p_{jk})$ of path $p_{jk}$ is defined as its precision:
\begin{equation}
	s(p_{jk}) = \frac{\mathbb{N}_{jk}}{\mathbb{M}_{jk}},
\end{equation}
where $\mathbb{M}_{jk}$ is the number of occurrences of $p_{jk}$ for the entities in the group and $\mathbb{N}_{jk}$ is the number of times that $p_{jk}$ reaches the correct tail entities of $r_j$. 

\subsubsection{Cross-type Reasoning Schema}
\textbf{The Motivation and Construction Process.} Some relations may appear in multiple entity type groups. For instance, the entity types of person and organization have common attributes (also treated as relations, as they are handled in the same way), such as $telephone\ number$ and $address$. Similarly, relations like $country$ are possessed by the entity types of person, organization, and geo-political entity, as exemplified in Figure~\ref{fig:model_example}. To handle this, we adopt an approach inspired by the bottom-up construction process of KG schema, where common relations of entity types are moved to their parent type. Analogically, we move the common relations of certain entity types to the cross-type group, which is considered as their parent group. As depicted in the left part of Figure~\ref{fig:model_framework}, relation $r_2$, appearing in all the four groups, is moved to the top cross-type group. Likewise, relation $r_3$, shared by person and organization, is also relocated. Illustratively, as demonstrated in Figure~\ref{fig:model_example}, $country$ possessed by person, organization, and geo-political entity is shifted to the top cross-type group. The corresponding paths are combined:
\begin{equation}
	P(r) = \cup_{i=1}^n P(r)_i,
\end{equation}
where $P(r)$ is the path set of the cross-type relation $r$, $\cup$ is the union operation, $n$ is the number of the entity type groups having $r$, and $P(r)_i$ is the path set of $r$ in its $i$-th entity type group.

\textbf{Path Score Calculation for the Cross-type Relations.}
The path scores of the cross-type relations are aggregated from the entity types where they appear. For a path $p$ of the cross-type relation $r$, its score is:
\begin{equation}
	s(p) = \frac{\sum_{i=1}^{n'} {\mathcal N}_i}{\sum_{i=1}^{n'} {\mathcal M}_i},
\end{equation}
where $n'$ is the number of the entity type groups in which $r$ and $p$ co-appear in their relation-path reasoning elements, ${\mathcal M}_i$ is the number of occurrences of $p$ for the $i$-th group, and ${\mathcal N}_i$ is the number of times that $p$ reaches the correct tail entities of $r$.

\subsubsection{Relation-path Reasoning Schema}
After shifting common relations to the cross-type group and aggregating their paths and scores, we obtain the final relation-path reasoning schema consisting of type-specific and cross-type ones. The paths of each relation are sorted in reverse order by their scores. Before sorting, these scores are adjusted based on the following consideration: Although for sparse KG, missing facts often make the model need to try more hops when short paths are unavailable, short paths are preferred. Actually, rules are typically short in length. Short paths are more likely to be rules. Therefore, we introduce a hop decay to the score of each path $p$:
\begin{equation}
	s(p) = f(len(p)) \cdot s(p),
	\label{eq:path_score}
\end{equation}
where $len(p)$ is the length of $p$, i.e., the number of hops of $p$, and $f(len(p))$ is defined as:
\begin{equation}
	f(len(p)) = d^{len(p)},
\end{equation}
where $d$ is the hop decay factor. The score is multiplied by $d$ for each increase in the number of hops.

Generally, the construction of the relation-path reasoning schema is a bottom-up approach. It thoroughly explores the training data, thereby alleviating the sparseness problem of sparse KGs. Besides, it enables us to obtain more precise path scores by considering paths from a global perspective.

\subsection{Stage2: Path Reasoning}
Based on the relation-path reasoning schema, we perform path reasoning and get the scores of candidate tail entities.

\subsubsection{The Reasoning Process}
In particular, we perform path reasoning as follows:
\begin{itemize}
	\item For each given query $(h,r,?)$, we retrieve the corresponding reasoning paths $[p_1, p_2, \cdots]$ by the entity type of $h$ and $r$.
	\item For the top $N_{top}$ paths, we start from $h$ and follow the paths to arrive at candidate tail entities. If a relation in the path does not appear, the process is halted midway. Here, $N_{top}$ is the number of explored top paths. We try top paths instead of all paths to relieve the impact from spurious paths.
	\item Sort all the candidate tail entities reversely by scores, which are introduced in the following.
\end{itemize}
\subsubsection{The Scores of Candidate Tail Entities}
We obtain the score of each candidate tail entity by summing the scores of the paths reaching it. The rationale is that individual indeterministic paths alone cannot be reliable. However, when multiple paths converge towards the same answer, it is possible to infer a meaningful result. Specifically, for a candidate tail entity $t_i$, its score $s(t_i)$ is:
\begin{equation}
	s(t_i) = \sum_{p\in Q} s(p),
\end{equation}
where $Q$ is the set of paths of relation $r$ that connect $h$ to $t_i$.

\subsubsection{One Update to the Scores}
The scores of candidate tail entities are further adjusted with the following answer similarity: The answer should be similar to other tail entities with the same $r$ and path. This is inspired by the phenomenon of analogy. It suggests that entities sharing the same $r$ and path may also exhibit similarities in their tail entities. For example, these tail entities may also belong to the same group.

Thus, the score $s(t_i)$ is multiplied by the similarity of the candidate and other similar tail entities:
\begin{equation}
	s(t_i) = \max_{t'\in T} cos\_sim (t_i, t')\cdot s(t_i),
	\label{eq:score}
\end{equation}
where $T$ is the set of other similar tail entities sharing the same $r$ and path, and $cos\_sim(t_i, t')$ is the cosine similarity between the vectors of $t_i$ and $t'$. The maximum similarity is used to mitigate the impact of spurious paths, which arrive at the answer coincidentally. To keep the explainability of LoGRe, we represent each entity as a $|R|$-hot vector, where $|R|$ is the size of the relation set $R$. An entry in the vector is set to 1 if the entity has that relation with some entities; otherwise, it is set to 0.

Then, the candidates are sorted by their scores and candidates with higher scores are more likely to be the correct tail entities.

\subsection{The Explainability of LoGRe}
The proposed LoGRe model is inherently explainable in its design. Moreover, for each given query, it outputs not only the candidate tail entities but also the reasoning paths with scores for each candidate tail entity. As illustrated in the right part of Figure~\ref{fig:model_framework}, for candidate tail entity $t_1$, the reasoning paths $p_2$ and $p_4$ with scores reveal the step-by-step processes from the head entity $h$ to $t_1$ via the relations.

\section{Experiments}
\label{sec:experiments}
We evaluate LoGRe on five benchmark sparse KG datasets and conduct comprehensive analyses to examine its performance.

\subsection{Datasets}
\label{subsec:datasets}
The adopted sparse KG datasets include FB15K-237-10\%, FB15K-237-20\%, FB15K-237-50\%, NELL23K, and WD-singer \cite{DacKGR}. The first three datasets randomly retain 10\%, 20\%, and 50\% facts from FB15K-237~\cite{FB15K-237}, respectively. WD-singer is a dataset of singer domain extracted from Wikidata~\cite{Wikidata}. NELL23K is a randomly sampled dataset from NELL~\cite{NELL}. The detailed statistics of these datasets are listed in Table~\ref{table:dataset}, where $|E|$, $|Train|$, $|Valid|$, and $|Test|$ are the sizes of the entity set, training set, validation set, and test set, respectively; Degree is the average degree of the entities in the corresponding training set.
\begin{table}[htb]
	\setlength{\tabcolsep}{0.43em}
	\centering
	\small
	\caption{The statistics of the five experimental datasets.}
	\begin{tabular}{c|c|c|c|c|c|c}
		\toprule[1pt]
		Dataset &$|E|$ &$|R|$ &$|Train|$ &$|Valid|$ &$|Test|$ &Degree\\
		\midrule[0.5pt]
		\!FB15K-237-10\% &11,512 &237 &27,211 &15,624 &18,150 &4.727\\
		\!FB15K-237-20\% &13,166 &237 &54,423 &16,963 &19,776 &8.267\\
		\!FB15K-237-50\% &14,149 &237 &136,057 &17,449 &20,324 &19.232\\
		NELL23K &22,925 &200 &25,445 &4,961 &4,952 &2.220\\
		WD-singer &10,282 &131 &15,906 &2,084 &2,134 &3.094\\
		\bottomrule[1pt]
	\end{tabular}
	\label{table:dataset}
\end{table}

\subsection{Experimental Settings}
\subsubsection{Baselines and Evaluation Metrics}
Baselines comprise three types of representative or state-of-the-art models. The first type is path-based models, of the same type with LoGRe, including DacKGR~\cite{DacKGR}, SparKGR~\cite{SparKGR}, DT4KGR~\cite{DT4KGR}, and Hi-KnowE~\cite{Hi-KnowE}. The second type is rule-based models\footnote{The latest rule-based model Ref.~\cite{Rule-SKGC-2023} for sparse KG completion is not applied as a baseline, as it necessitates an additional source KG for rule transfer.}, including NTP~\cite{NTP}, RlvlR~\cite{RlvlR}, and AnyBURL~\cite{AnyBURL}.  We also refer to embedding-based models, including TuckER~\cite{TuckER}, ConvE~\cite{ConvE}, InteractE~\cite{InteractE}, NBFNet~\cite{NBFNet}, KRACL~\cite{SKGC2023-KRACL}, and Ref.~\cite{SKG2023-Application}. Lastly, we compare to ChatGPT\footnote{https://openai.com/blog/chatgpt/}, the prominent large language model.

For every fact $(h, r, t)$ in the test set, we convert it to a query $(h, r, ?)$ and apply models to get the ranking list of candidate tail entities, following Refs.~\cite{DacKGR,SparKGR}. The standard Mean Reciprocal Rank (MRR) and Hits@K are adopted as evaluation metrics for model comparison. MRR is the mean reciprocal rank of all the correct tail entities among the candidates and Hits@K is the proportion of correct tail entities ranking within the top K positions. Higher values of MRR and Hits@K indicate better performance.

\begin{table*}[htb]
	\setlength{\tabcolsep}{0.27em}
	\centering
	\small
	\caption{Experimental results in terms of MRR and Hits@\{3, 10\} (\%). The best scores of rule-based (the second block) and path-based models (the third block) are in bold, and the best scores of embedding-based models (the first block) are underlined.}
	\makebox[\columnwidth][c]{
		\begin{tabular}{c|ccc|ccc|ccc|ccc|ccc}
			\toprule[1pt]
			\multirow{2}{*}{Model} &\multicolumn{3}{c|}{FB15K-237-10\%} &\multicolumn{3}{c|}{FB15K-237-20\%} &\multicolumn{3}{c|}{FB15K-237-50\%} &\multicolumn{3}{c|}{NELL23K} &\multicolumn{3}{c}{WD-singer}\\
			\cline{2-16} &MRR &Hits@3 &Hits@10 &MRR &Hits@3 &Hits@10 &MRR &Hits@3 &Hits@10 &MRR &Hits@3 &Hits@10 &MRR &Hits@3 &Hits@10\!\\
			\midrule[0.5pt]
			TuckER &0.252 &27.2 &\underline{40.4} &\underline{0.268} &\underline{28.9} &\underline{42.8} &0.314 &34.2 &50.1 &\underline{0.276} &\underline{30.2 }&46.7 &0.421 &47.1 &57.1\\
			ConvE &0.245 &26.2 &39.1 &0.261 &28.3 &41.8 &0.313 &34.2 &50.1 &\underline{0.276} &30.1 &46.4 &0.448 &47.8 &56.9\\
			InteractE &0.254 &27.3 &40.2 &0.265 &28.8 &\underline{42.8} &0.308 &33.6 &49.7 &0.265 &28.6 &46.7 &0.435 &47.6 &57.4\\
			NBFNet &0.241 &26.3 &38.8 &0.260 &27.8 &41.7 &\underline{0.316} &34.1 &50.3 &0.274 &28.9 &46.9 &\underline{0.453} &\underline{49.3} &\underline{58.9}\\
			KRACL &0.164 &17.0 &21.2 &0.170 &16.9 &19.8 &0.222 &26.8 &44.4 &0.158 &15.8 &27.6 &0.142 &13.4 &20.7\\  
			Ref.~\cite{SKG2023-Application} &\underline{0.268} &\underline{27.9} &38.7 &0.267 &28.2 &42.1 &0.309 &\underline{35.0} &\underline{51.9} &0.261 &29.6 &\underline{47.2} &- &- &-\\
			\midrule[0.5pt]
			NTP &0.083 &11.4 &16.9 &0.173 &16.1 &21.7 &0.222 &23.1 &30.7 &0.132 &14.9 &24.1 &0.292 &31.1 &44.2\\
			RlvlR &0.107 &12.2 &20.6 &0.132 &15.2 &27.1 &0.199 &20.8 &32.4 &0.152 &17.3 &25.0 &0.374 &32.0 &47.6\\
			AnyBURL &0.149 &15.5 &26.7 &0.164 &16.7 &29.3 &0.198 &21.3 &35.1 &0.176 &18.5 &25.2 &0.392 &34.1 &48.6\\
			\midrule[0.5pt]
			DacKGR &0.218 &23.9 &33.7 &0.242 &27.2 &38.9 &0.293 &32.0 &45.7 &0.197 &20.0 &31.6 &0.377 &42.1 &48.5\\
			SparKGR &\textbf{0.228} &24.5 &35.0 &0.252 &27.7 &39.1 &0.292 &32.0 &46.2 &0.203 &22.2 &33.9 &0.393 &43.7 &50.7\\
			DT4KGR &- &- &- &0.254 &- &40.1 &\textbf{0.297} &- &46.2 &- &- &- &- &-\\
			Hi-KnowE &0.224 &\textbf{25.5} &34.1 &0.247 &27.7 &38.1 &- &- &- &- &- &- &- &- &-\\
			\midrule[0.5pt]
			LoGRe &\textbf{0.228} &24.5 &\textbf{36.2} &\textbf{0.261} &\textbf{28.0} &\textbf{41.3} &\textbf{0.297} &\textbf{32.7} &\textbf{46.4} &\textbf{0.259} &\textbf{27.9} &\textbf{41.7} &\textbf{0.459} &\textbf{48.9} &\textbf{54.5}\\
			\bottomrule[1pt]
		\end{tabular}
	}
	\label{table:SKGC}
\end{table*}

\begin{table}[htb]
	\centering
	\small
	\caption{Comparison with ChatGPT on Hits@1 (\%).}
	\begin{tabular}{c|c|c|c}
		\toprule[1pt]
		Model &FB15K-237-50\% &NELL23K &WD-singer\\
		\midrule[0.5pt]
		ChatGPT &14.0 &10.8 &32.7\\
		LoGRe &\textbf{21.2} &\textbf{18.4} &\textbf{40.6}\\
		\bottomrule[1pt]
	\end{tabular}
	\label{table:ChatGPT}
\end{table}
\subsubsection{Implementation Details}
\label{subsubsec:implementation_details}
\textbf{Entity Types.} LoGRe requires entity types\footnote{If all entity types are unavailable to obtain, entities can be grouped using various clustering algorithms, such as hierarchical clustering, where entities sharing numerous relations are clustered into a group.} to construct the relation-path reasoning schema and perform path reasoning. For the FB15K-237 series, we retrieve entity types from FB15K-237~\cite{TKRL}. For NELL23K, entity types are obtained from entity names, as they are in the form of ``concept\_type\_entity''. For WD-singer, we obtain entity types from Wikidata~\cite{Wikidata} via the tail entities of relation ``$instance\ of$''. When there are multiple types, we select the most frequent one\footnote{For the FB15K-237 series, ``/common/topic'' is removed before selection, as almost all entities belong to this type.} as the type to mitigate the impact of noisy data. If no type is available for an entity, it is assigned to the type ``others''.

\textbf{Hyper-parameters.} The hyper-parameters of LoGRe are selected from the following ranges in terms of MRR: The maximum number of collected paths $N_{path}$ $\in$ $\{1000,$ $5000,$ $10000,$ $15000,$ $20000\}$, the maximum number of path hops $N_{hop}\in \{3,$ $4,$ $5,$ $6\}$, the number of explored top paths $N_{top} \in \{100, 500, 1000\}$, and the hop decay factor $d\in \{0.95, 0.9, 0.8, 0.7, \cdots, 0.1\}$. The applied settings are: $N_{path} = 20000$, $N_{hop} = 6$, $N_{top} = 1000$, $d = 0.95$ for FB15K-237-10\%, $N_{path} = 5000$, $N_{hop} = 5$, $N_{top} = 500$, $d = 0.6$ for FB15K-237-20\%, $N_{path} = 1000$, $N_{hop} = 4$, $N_{top} = 100$, $d = 0.8$ for FB15K-237-50\%, $N_{path} = 10000$, $N_{hop} = 6$, $N_{top} = 100$, $d = 0.5$ for NELL23K, and $N_{path} = 20000$, $N_{hop} = 6$, $N_{top} = 100$, $d = 0.2$ for WD-singer.

\textbf{Baseline Results.} Baseline results reported in \cite{SparKGR} are used, while the results of Ref.~\cite{SKG2023-Application}, DT4KGR~\cite{DT4KGR}, and Hi-KnowE~\cite{Hi-KnowE}\footnote{DT4KGR and Hi-KnowE are available online on 19 January 2024 and 11 April 2024, respectively, lacking source codes, and with certain implementation details undisclosed. Thus, some of their results are missing in Table~\ref{table:SKGC}.} are obtained from the original papers. We further tune the hyper-parameters of KRACL~\cite{SKGC2023-KRACL} on the five experimental datasets to get the best results. For ChatGPT, we follow the state-of-the-art LambdaKG~\cite{LambdaKG} to construct the prompt, which comprises the task description with candidates, demonstrations, and test information:
\begin{itemize}
	\item The task description is: ``Given head entity and relation, predict the tail entity from the following candidates or others: $C$''. Here, $C$ is the top 100 most relevant entities from the training set, obtained by the retrieval model BM25~\cite{BM25}.
	\item Similarly, the demonstrations utilize the top 5 most similar training data as examples.
	\item The test information is ``What is the $r$ of $h$?'', given the query $(h, r, ?)$.
\end{itemize}
During this process, the names of each entity and relation are required. We get the entity names and relation names of the FB15K-237 series from LambdaKG~\cite{LambdaKG}. Those of NELL23K are obtained by removing the ``concept'' prefix and segmenting the remaining characters, while those of WD-singer are retrieved from Wikidata~\cite{Wikidata}. For a fair comparison, ChatGPT solely utilizes these fundamental entity and relation names, without any supplementary information.

\subsection{Results of Sparse KG Completion}
\label{subsec:SKGC_results}
The experimental results are presented in Tables~\ref{table:SKGC} and \ref{table:ChatGPT}. For the experiments on ChatGPT, we focus on FB15K-237-50\%, which has the largest test set among the FB15K-237 series, NELL23K, and WD-singer. As we can only get Hits@1 by evaluating the accuracy of the responses from ChatGPT, we adopt only Hits@1 following LambdaKG~\cite{LambdaKG}. For the other baselines, we utilize all the five datasets and follow Refs.~\cite{DacKGR,SparKGR,SKG2023-Application} to apply MRR and Hits@\{3, 10\} as metrics. From the results, we have the following observations:

\textbf{Compared with rule-based and path-based models,} LoGRe outperforms them (except for Hi-KnowE~\cite{Hi-KnowE}), particularly on the challenging dataset NELL23K, where the state-of-the-art baselines gain the worst performance among the five datasets. LoGRe significantly improves all the metrics on NELL23K, 0.056 on MRR (27.6\% relative improvement), 5.70\% on Hits@3 (25.7\% relative improvement), and 7.80\% on Hits@10 (23.0\% relative improvement). LoGRe is better or comparable than much more intricate Hi-KnowE~\cite{Hi-KnowE}, which relies on AnyBURL~\cite{AnyBURL}, ConvE~\cite{ConvE}, and Transformer~\cite{Transformer}. This validates the effectiveness of LoGRe looking inward globally. Additionally, state-of-the-art path-based models perform better than state-of-the-art rule-based ones. The poor performance of rule-based models may be due to the large number of truncated paths in sparse KG, which prevents these models from effective rule mining. Whereas, path-based models have more flexibility in exploring KG rather than relying on strict rules. This enables them to consider various possibilities for reaching the correct tail entities. LoGRe further aggregates and evaluates the paths and candidate tail entities from a global perspective. Thus, LoGRe is more capable of helping the head entities select more high-quality paths to arrive at the correct tail entities than path-based baselines.

\textbf{LoGRe is comparable but worse than embedding-based models.} However, embedding-based models serve as a reference only, as they are different categories of models as opposed to our path-based model. Unlike path-based models, embedding-based models usually require tuning many more parameters carefully and are not explainable. The same phenomenon that embedding-based models outperform rule-based and path-based models is revealed in previous path-based studies~\cite{DacKGR,SparKGR}. The reason is that embedding-based models implicitly encoding the facts of KG into vectors, are free from rule mining and path traversing. Thus, they are less affected by path truncation in sparse KG. Combining the superior performance of embedding-based models and the explainability of path-based models is our future work.

\textbf{Compared with ChatGPT,} LoGRe performs better by more than 7.0\% on all the three datasets. We attribute this to the difficulty of sparse KG completion. As displayed in Table~\ref{table:ave_hops}, we often need to explore multiple hops to get the correct tail entities on sparse KG. However, ChatGPT, targeting natural language understanding, cannot well handle structure and path reasoning. Thus, it is challenging for ChatGPT to get accurate results via limited context understanding. In contrast, LoGRe excels in multi-hop exploration. 

\begin{table}[htb]
	\setlength{\tabcolsep}{0.45em}
	\centering
	\small
	\caption{Average hops to the correct tail entities on test sets. FB15K-237 is denoted as FB.}
	\begin{tabular}{c|c|c|c|c|c}
		\toprule[1pt]
		Dataset &FB-10\% &FB-20\% &FB-50\% &NELL23K &WD-singer\\
		\midrule[0.5pt]
		Average hops &5.372 &4.603 &3.493 &4.436 &4.093\\
		\bottomrule[1pt]
	\end{tabular}
	\label{table:ave_hops}
\end{table}

\subsection{Ablation Study}
LoGRe has two stages, with the construction of the type-specific reasoning schema in stage 1 and stage 2 (path reasoning) being essential. Thus, we remove the construction of the cross-type reasoning schema in stage 1 to evaluate its effectiveness. This ablation is denoted as LoGRe(-cross\_type). Besides, LoGRe prefers short ones via hop decay and considers the answer similarity. To study their contributions to the performance, we remove one of them each time, denoted as LoGRe(-prefer\_shortpath) and LoGRe(-answer\_similarity), respectively. The experimental results are reported in Table~\ref{table:ablation_study}.

\begin{table*}[htb]
	\setlength{\tabcolsep}{0.19em}
	\centering
	\small
	\caption{Ablation study results regarding MRR and Hits@\{3, 10\} (\%).}
	\makebox[\columnwidth][c]{
		\begin{tabular}{c|ccc|ccc|ccc|ccc|ccc}
			\toprule[1pt]
			\multirow{2}{*}{Model} &\multicolumn{3}{c|}{FB15K-237-10\%} &\multicolumn{3}{c|}{FB15K-237-20\%} &\multicolumn{3}{c|}{FB15K-237-50\%} &\multicolumn{3}{c|}{NELL23K} &\multicolumn{3}{c}{WD-singer}\\
			\cline{2-16} &MRR &Hits@3 &Hits@10 &MRR &Hits@3 &Hits@10 &MRR &Hits@3 &Hits@10 &MRR &Hits@3 &Hits@10 &MRR &Hits@3 &Hits@10\\
			\midrule[0.5pt]
			LoGRe(-cross\_type) &0.221 &23.7 &35.1 &0.257 &27.6 &40.9 &0.294 &32.4 &46.3 &0.219 &23.2 &36.1 &0.448 &48.1 &53.7\\
			LoGRe(-prefer\_shortpath) &\textbf{0.228} &24.4 &36.1 &\textbf{0.261} &\textbf{28.0} &41.2 &0.295 &\textbf{32.7} &46.3 &0.254 &27.7 &41.3 &0.415 &46.0 &53.4\\
			LoGRe(-answer\_similarity) &0.227 &24.3 &35.9 &0.260 &27.8 &41.1 &0.294 &32.3 &46.0 &0.255 &27.3 &41.5 &0.455 &\textbf{48.9} &\textbf{54.5}\\
			LoGRe &\textbf{0.228} &\textbf{24.5} &\textbf{36.2} &\textbf{0.261} &\textbf{28.0} &\textbf{41.3} &\textbf{0.297} &\textbf{32.7} &\textbf{46.4} &\textbf{0.259} &\textbf{27.9} &\textbf{41.7} &\textbf{0.459} &\textbf{48.9} &\textbf{54.5}\\
			\bottomrule[1pt]
		\end{tabular}
	}
	\label{table:ablation_study}
\end{table*}

Based on the results, we observe that removing any of them reduces the effectiveness of LoGRe, which confirms the necessity of all the three designs. Particularly, removing the construction of the cross-type reasoning schema significantly decreases the performance. It underscores the importance of shifting common relations to the cross-type group and aggregating their paths and scores globally. Additionally, not giving preference to short paths on WD-singer leads to a more obvious drop in performance compared to the other datasets. It is reasonable, as WD-singer is specific to the singer domain, while the other datasets contain general facts. Specific WD-singer may have simpler relation patterns among entities, and when there are high-quality short paths, long paths are less important. This observation aligns with the small optimal hop decay factor, i,e, 0.2, as introduced in Section~\ref{subsubsec:implementation_details}. In contrast, on general-domain datasets like the FB15K-237 series, long paths are crucial for reasoning out the answer across entities of diverse types.

\subsection{Case Study}
To illustrate the explainability of LoGRe, we conduct case study on WD-singer, following DacKGR~\cite{DacKGR}. Two representative examples, $(Louise\ Kirkby\ Lunn, citizenship,$ $?)$ and $(Carlo\ Scattola,$ $participant\ in, ?)$ are selected from the test set. The first one is about person's basic information, also selected in DacKGR~\cite{DacKGR} and the second one relates to person's activities. We find that the corresponding correct tail entities $United\ Kingdom$ and $La\ Cenerentola\ o\ sia\ La$ $virt\grave{u}\ in\ trionfo$ are ranked first by LoGRe among the candidates. The used reasoning paths are presented in Table~\ref{table:case_study}. For space limitation, we present the top 10 high-score paths.
\begin{table}[htb]
	\setlength{\tabcolsep}{0.15em}
	\centering
	\small
	\caption{The top 10 high-score paths of LoGRe on two representative cases. Order$'$ is the path order without hop decay and $r^{-1}$ is the inverse relation of $r$. Case 2 only has 3 paths.}
	\begin{tabular}{c|c|c}
		\toprule[1pt]
		\multicolumn{3}{c}{Query: $(Louise\ Kirkby\ Lunn, citizenship, ?)$}\\
		\midrule[0.5pt]
		Order &Path &\makecell[c]{Order$'$}\\
		\midrule[0.5pt]
		1 &$(student^{-1}, student, citizenship)$ &1\\
		2 &$(student\ of, student\ of^{-1}, citizenship)$ &3\\
		3 &$(student\ of, student, citizenship)$ &4\\
		4 &$(student^{-1}, student\ of^{-1}, citizenship)$ &5\\
		\midrule[0.5pt]
		5 &\makecell[c]{$(student^{-1}, student, citizenship, citizenship^{-1},$\\ $citizenship)$} &2\\
		6 &\makecell[c]{$(student^{-1}, student\ of^{-1}, citizenship, citizenship^{-1},$\\ $citizenship)$} &6\\
		7 &\makecell[c]{$(student\ of, student\ of^{-1}, citizenship, citizenship^{-1},$\\ $ citizenship)$} &7\\
		8 &\makecell[c]{$(student\ of, student, citizenship, citizenship^{-1},$\\ $ citizenship)$} &8\\
		\midrule[0.5pt]
		9 &\makecell[c]{$(student\ of, student\ of^{-1}, student\ of, student\ of^{-1},$\\ $ citizenship)$} &9\\
		10 &\makecell[c]{$(student\ of, student, student\ of, student\ of^{-1},$\\ $ citizenship)$} &10\\
		\bottomrule[1pt]
		\toprule[1pt]
		\multicolumn{3}{c}{Query: $(Carlo\ Scattola, participant\ in, ?)$}\\
		\midrule[0.5pt]
		1 &$(cast\ member^{-1})$ &3\\
		2 &$(cast\ member^{-1}, cast\ member, participant\ in)$ &1\\
		3 &\makecell[c]{$(cast\ member^{-1}, cast\ member, cast\ member^{-1},$\\ $ cast\ member, participant\ in)$} &2\\
		\bottomrule[1pt]
	\end{tabular}
	\label{table:case_study}
\end{table}

From the upper part of Table~\ref{table:case_study}, it can be observed that the 5-hop path $(student^{-1}, student,$ $citizenship, citizenship^{-1},$ $citizenship)$ ranks second when hop decay is disabled; otherwise, its ranking drops to more appropriate fifth place. This also proves the necessity of our preferring short paths. Furthermore, as $student^{-1}$ is equal to $student\ of$ and $student\ of^{-1}$ is equal to $student$, the first four paths tell that the $citizenship$ of a person may be the same as that of his/her classmate; the middle four paths suggest that it may be the same as the $citizenship$ of someone who shares the $citizenship$ with his/her classmate; the remaining two paths imply that it may be the same as the $citizenship$ of his/her classmate's classmate. Interestingly, LoGRe finds weak signals like the above ones and gathers them to determine the answer. In contrast, existing path-based studies typically use one path to get the answer, extremely difficult and indeterministic. Actually, the above indeterministic paths alone are not worthy of attention, while if many indeterministic paths point to the same answer, it makes sense.

From the bottom part of Table~\ref{table:case_study}, we have the same observation. LoGRe uses all available signals, including the deterministic signal that $participant\ in$ is equal to $cast\ member^{-1}$ and the indeterministic signals that a person may be $participant\ in$ something in which his/her co-participant participates and his/her co-participant's co-participant participates, to obtain a relatively deterministic answer.

\subsection{Hyper-parameter Analysis}
\begin{figure*}[!htb]
	\centering
	\includegraphics[width=6.95in]{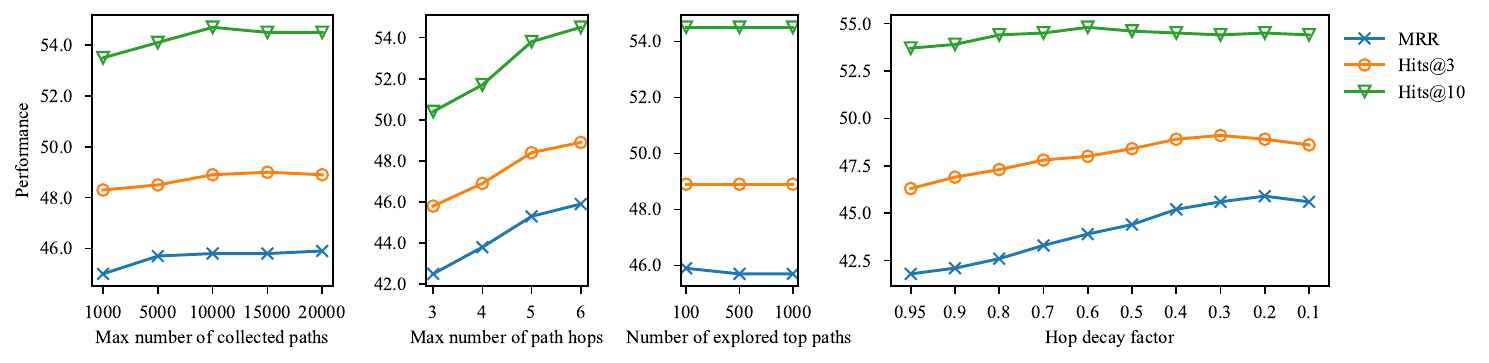}  
	\caption{Hyper-parameter analysis of LoGRe on WD-singer.}
	\label{fig:hyperparameter_analysis}
\end{figure*}

The hyper-parameters of LoGRe are the maximum number of collected paths, maximum number of path hops, number of explored top paths, and hop decay factor. We thoroughly analyze these hyper-parameters to evaluate their impacts on the performance. Without loss of generality, we also conduct the analysis on WD-singer. The results are exhibited in Figure~\ref{fig:hyperparameter_analysis}.

\begin{table*}[htb]
	\setlength{\tabcolsep}{1em}
	\centering
	\small
	\caption{Complexity comparison with the best baselines. Here, $\alpha$ is the embedding dimension, $\beta$ is the number of parameters in the neural networks, $N_{epoch}$ is the number of training epochs, and $\mathcal{S} = N_{time}N_{hop}$ and $\mathcal{T} = N_{time}N_{hop} + (|Valid| + |Test|) N_{rule} N_{hop}$ are the space and time complexities of the rule mining algorithm AnyBURL~\cite{AnyBURL}, respectively. $N_{time}$ is the number of times for the path exploration and $N_{rule}$ is the number of matched rules for each query.}
	\begin{tabular}{c|c|c}
		\toprule[1pt]
		Model &Space complexity &Time complexity\\
		\midrule[0.5pt]
		TuckER &$O(|E|\alpha+|R|\alpha+\alpha^3)$ &$O((N_{epoch} |Train| + |Valid| + |Test|)\alpha^3)$\\  
		ConvE &$O(|E|\alpha+|R|\alpha+{\beta})$ &$O((N_{epoch} |Train| + |Valid| + |Test|) \alpha^2)$\\  
		InteractE &$O(|E|\alpha+|R|\alpha+{\beta})$ &$O((N_{epoch} |Train| + |Valid| + |Test|) \alpha^2)$\\  
		NBFNet &$O(|R|\alpha^2+\beta)$ &$O((N_{epoch} |Train| + |Valid| + |Test|) \alpha^2)$\\
		SparKGR &$O(\mathcal{S} +|E|\alpha+|R|\alpha+{\beta})$ &$O(\mathcal{T} + (N_{epoch} |Train| + |Valid| + |Test|) \alpha^2)$\\
		DT4KGR &$O(\mathcal{S}+|E|\alpha+|R|\alpha+{\beta})$ &$O(\mathcal{T} + (N_{epoch} |Train| + |Valid| + |Test|) \alpha^2)$\\
		Hi-KnowE &$O(\mathcal{S}+|E|\alpha+|R|\alpha+{\beta})$ &$O(\mathcal{T} + (N_{epoch} |Train| + |Valid| + |Test|) \alpha^2)$ \\
		LoGRe &$O(|E| N_{path} N_{hop})$ &$O(|E| N_{path} N_{hop} + (|Valid| + |Test|) N_{top} N_{hop})$\\
		\bottomrule[1pt]
	\end{tabular}
	\label{table:complexity}
\end{table*}

It can be observed from the first and third subfigures of Figure~\ref{fig:hyperparameter_analysis} that the maximum number of collected paths and number of explored top paths impact the performance of WD-singer slightly. Notably, 20000 and 100, respectively, yield marginally superior results on MRR. Concerning the second subfigure, the performance improves on all the metrics as the maximum number of path hops increases. When the maximum number of path hops is 6, LoGRe achieves the best performance. As illustrated in the fourth subfigure, a decrease in the hop decay factor leads to performance improvement. The optimal value is 0.2, indicating a substantial penalization of long paths. Overall, the performance variations of different parameter values are limited, demonstrating the robustness of LoGRe.

\subsection{Complexity Analyses} 
Model complexity comprises space and time aspects. We compare LoGRe with the best baselines\footnote{Ref.~\cite{SKG2023-Application} is not included in the comparison, as it introduces external knowledge graphs and an entity linking pipeline, rendering it significantly more complex.} for these two dimensions. In terms of space complexity, LoGRe only necessitates storage for the relation-path reasoning schema of space complexity $O(|E| N_{path} N_{hop})$. As presented in Table~\ref{table:complexity}, it generally occupies more space compared to excellent embedding-based KG completion models TuckER~\cite{TuckER}, ConvE~\cite{ConvE}, InteractE~\cite{InteractE}, and NBFNet~\cite{NBFNet}. However, it stands out as the most space-efficient option when compared to the state-of-the-art sparse KG completion models SparKGR~\cite{SparKGR}, DT4KGR~\cite{DT4KGR}, and Hi-KnowE~\cite{Hi-KnowE}, which rely on external assistance. These models typically require embeddings for all entities and relations, parameters for neural networks, and storage space for rules.

Regarding time complexity, LoGRe does not require training. It mainly involves constructing the relation-path reasoning schema, which exhibits a time complexity of $O(|E| N_{path} N_{hop})$, and conducting path reasoning on the validation or test set of time complexity $O(|Valid| N_{top} N_{hop})$ or $O(|Test| N_{top} N_{hop})$. The time complexity associated with the preference for short paths and answer similarity is considerably smaller and can be disregarded. Thus, the overall time complexity is $O(|E| N_{path} N_{hop} + (|Valid| + |Test|) N_{top} N_{hop})$. In contrast, the best baselines necessitate training over numerous epochs. As illustrated in Table~\ref{table:complexity}, their complexity can be summarized as at least $O((N_{epoch} |Train| + (|Valid| + |Test|) g(\alpha))$ (some models need additional time for rule mining). Here, $g(\alpha)$ is the complexity of the embedding-based networks. Given that typically $N_{epoch}|Train| > N_{path}N_{hop}$, $g(\alpha) > |E|$, and $g(\alpha) > N_{top} N_{hop}$, LoGRe is generally more time-efficient than existing KG completion models, especially in comparison to the three sparse KG completion models SparKGR~\cite{SparKGR}, DT4KGR~\cite{DT4KGR}, and Hi-KnowE~\cite{Hi-KnowE}, which require additional time for rule mining. Notably, we do not compare the actual run times of these methods, as run times significantly influenced by the configuration of the realistic computing environment. Instead, we compare their time complexities, which are more objective and independent of the environment.

\section{Conclusion and Future Work}
\label{sec:conclusion_futurework}
In this paper, we looked inward rather than seeking external help for the first time in the line of path-based sparse KG completion models. We proposed a two-stage model LoGRe, which constructs a relation-path reasoning schema by globally analyzing the training data to alleviate the sparseness problem and then aggregates paths to reason out answers. Experiments on five benchmark sparse KG datasets demonstrate that LoGRe outperforms rule-based and path-based baselines. It is also comparable to the referred embedding-based models and significantly superior to ChatGPT. Further comprehensive analyses substantiate the effectiveness and efficiency of LoGRe.

This paper looks back to the path-based research line and proposes an explainable model, in contrast to the prevailing trend of embedding-based models and large language models. We hope to inspire research endeavors that prioritize explainability, lacking in the current research landscape. In the future, we intend to enhance LoGRe by introducing external models and knowledge, such as harnessing the natural language understanding capabilities of large language models as mentioned in Section~\ref{subsec:SKGC_results}, while striving to preserve a high degree of explainability.

\begin{acks}
The work is supported by the SMP-IDATA Open Youth Fund, the Lenovo-CAS Joint Lab Youth Scientist Project, the project under Grant No. JCKY2022130C039, the Strategic Priority Research Program of the CAS under Grant No. XDB0680102, the National Natural Science Foundation of China under Grant No. 62002341, and the JCJQ Project of China.
\end{acks}

\bibliographystyle{ACM-Reference-Format}
\balance
\bibliography{CIKM2024}


\end{document}